\pdfoutput=1
\documentclass[runningheads]{llncs}
\usepackage{graphicx}

\usepackage{tikz}
\usepackage{comment}
\usepackage{amsmath,amssymb} 
\usepackage{color}
\usepackage[accsupp]{axessibility}  

\usepackage[width=122mm,left=12mm,paperwidth=146mm,height=193mm,top=12mm,paperheight=217mm]{geometry}

\begin{document}
\pagestyle{headings}
\mainmatter
\def\ECCVSubNumber{3117}  

\title{GenText: Unsupervised Artistic Text Generation via Decoupled Font and Texture Manipulation} 

\titlerunning{ } 
\authorrunning{ }
\author{Qirui Huang\inst{1} \and
Bin Fu\inst{1} \and
Aozhong zhang\inst{1} \and
Yu Qiao\inst{1}}
\institute{Shenzhen Institutes of Advanced Technology, Chinese Academy of Sciences}

\maketitle

\begin{abstract}
Automatic artistic text generation is an emerging topic which receives increasing attention due to its wide applications. 
The artistic text can be divided into three components,  content, font, and texture, respectively. 
Existing artistic text generation models usually focus on manipulating one aspect of the above components, which is a sub-optimal solution for controllable general artistic text generation. 
To remedy this issue, we propose a novel approach, namely GenText, to achieve general artistic text style transfer by separably migrating the font and texture styles from the different source images to the target images in an unsupervised manner. 
Specifically, our current work incorporates three different stages, stylization, destylization, and font transfer, respectively, into a unified platform with a single powerful encoder network and two separate style generator networks, one for font transfer, the other for stylization and destylization. 
The destylization stage first extracts the font style of the font reference image, then the font transfer stage generates the target content with the desired font style. 
Finally, the stylization stage renders the resulted font image with respect to the texture style in the reference image. 
Moreover, considering the difficult data acquisition of paired artistic text images, our model is designed under the unsupervised setting, where all stages can be effectively optimized from unpaired data.
Qualitative and quantitative results are performed on artistic text benchmarks, which demonstrate the superior performance of our proposed model. 
The code with models will become publicly available in the future.

\keywords{artistic text generation, unsupervised learning, generative adversarial networks}

\end{abstract}

\section{Introduction}
\label{sec:intro}



With the advances of image style transfer, artistic text generation has become an important image creation task in the computer vision community. 
The artistic text style transfer task aims to render the style of texts according to the reference images, which has a wide demand in various artistic creation applications. 
The challenge of the artistic text style transfer task mainly comes from the large diversity of artistic text and the difficulty of paired-data acquisition.

\begin{figure}[t]
\centering
\includegraphics[scale=0.25]{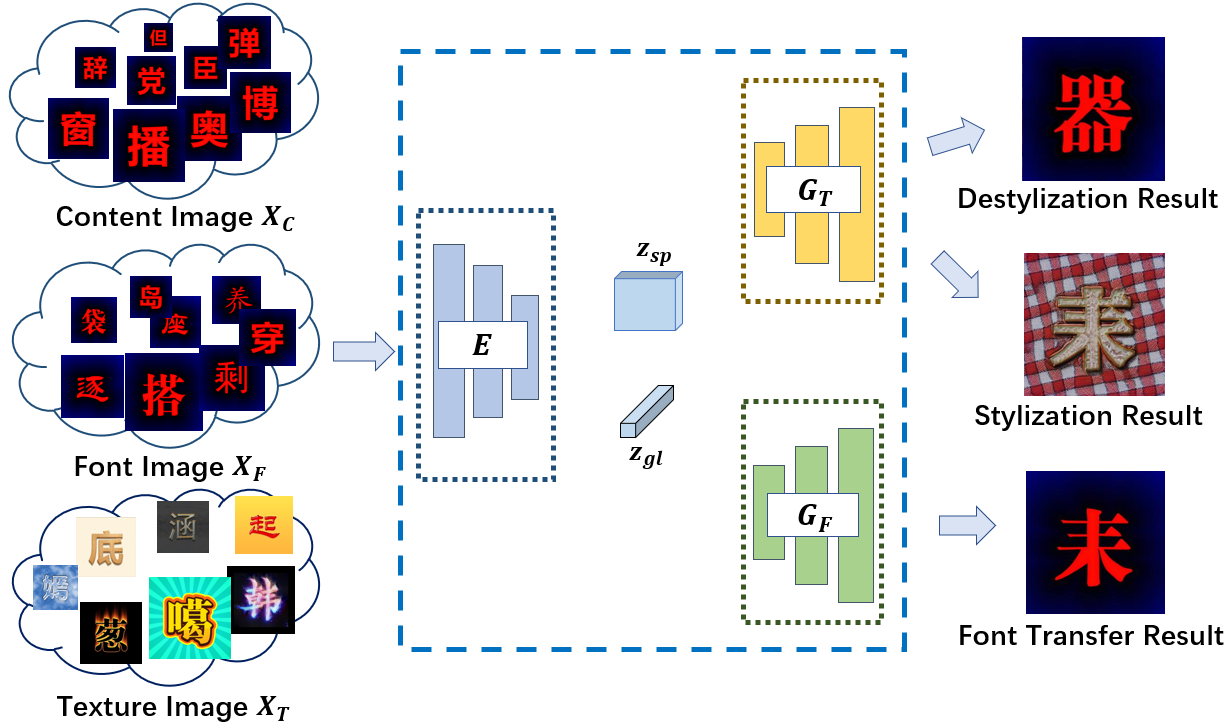}
\vspace{-13pt}
\caption{The overview of our proposed method. The artistic text can be divided into three components, including content, font, and texture. The encoder network $E$ project above information into the spatial code $z_{sp}$ and global code $z_{gl}$. Two style generator networks are employed to perform artistic text generation tasks, one ($G_F$) for font transfer, the other ($G_T$) for stylization and destylization.}
\label{fig:overview}
\vspace{-18pt}
\end{figure} 

Diversity is the essential characteristic of artistic text, which can be roughly sorted into three categories, content diversity, font diversity, and texture diversity, respectively. 
As shown in Fig. \ref{fig:overview}, the content diversity is referred to the intrinsic differences between different characters, which sometimes shows a significant variance in some language systems such as Chinese and Japanese. 
For specific content, the artistic style can be characterized by the font style and texture style, respectively. 
However, existing artistic text generation models usually focus on manipulating one aspect of the above components.
For example, on one hand, \cite{dgfont} pays attention {to} {the} font generation sub-task via employing deformable convolution to transfer font style from the reference image.
On the other hand, \cite{yang2019tet} focuses on visual effect transfer and {renders} the font image in the style specified by the artistic text. 
The above approaches only perform one kind of artistic text rendering {tasks}, namely font transfer or visual effect transfer, which significantly restrict visual creation in {real-world} applications. 

The usage of {paired data} in {the} optimization process is another serious drawback of recent models. 
As the data-driven nature of {the} deep network, mass data is {demanded} for optimizing a robust style transfer model and learning a rich style {representation} of {the} artistic text. 
Although there are abundant artistic texts in {the} real world, most of them are unavailable for optimizing existing generation models due to the requirement of {paired image} in {a} supervised setting. 
Therefore, the unsupervised {learning-based} style transfer method is a more plausible solution for artistic text generation {tasks}. 

To remedy {the} above issues, we propose a general artistic text generation model, namely GenText, to achieve fine adjustment of {the} artistic text generating process through the corresponding reference images, where the term “general" refers {to} two important advantages of our model. 
On one hand, we regard the content, font style, and texture style as three basic components to construct the integrated representation for artistic text, and thus our model can perform more general artistic text creation tasks.
To be concrete, we utilize a unified deep encoder network to encode the detailed structure relations into a spatial code, while encode the font and texture characteristics into a global vector. 
Two style generator networks are further proposed to perform artistic text generation tasks based on global style vector, one for font transfer, the other for stylization and destylization. 
Based on the different functions, our model can be separated into three different stages. 
The destylization stage first extracts the font style of the font reference image, then the font transfer stage generates the target content with the desired font style. 
Finally, the stylization stage renders the resulted font image with respect to the texture style in {the} reference image. 
Therefore, our model can not only perform traditional font transfer and artistic effect rendering tasks but also achieve the flexible adjustment of the font and texture styles in arbitrary text. 
On the other hand, our method is designed in {an} unsupervised setting, where all artistic texts can be effectively utilized to optimize a robust generation model with rich styles. 

To validate the effectiveness of our proposed method, we implement our model to perform the ablation study on a recently released artistic text benchmark \cite{te141k}. 
For traditional artistic effect rendering task, our model achieves the best record compared with the unsupervised image style transfer models. 
In the artistic text flexible adjustment setting, our proposed approach brings a significant improvement in image quality. 
Moreover, we further utilize our model to perform {open-set} artistic text generation to validate the generalization ability of our method. 

In summary, our contributions are threefold:
\begin{itemize}
\item We propose a general artistic text generation method, namely GenText, to perform flexible adjustment by generating an artistic character defined by content images with the font and texture styles specified by different reference images. 
\item To remedy the difficulty of paired-data acquisition, we optimize our GenText in unsupervised setting, where all artistic texts can be effectively utilized in our method. 
\item Experimental results demonstrate the effectiveness of our GenText. For the stylization task, our model achieves {the} best record in {unsupervised-based} image style transfer methods. For the flexible adjustment of artistic text, our GenText produces {high-quality} artistic texts compared with {the} baseline model. 
\end{itemize}

\begin{figure*}[t]
\centering
\includegraphics[scale=0.12]{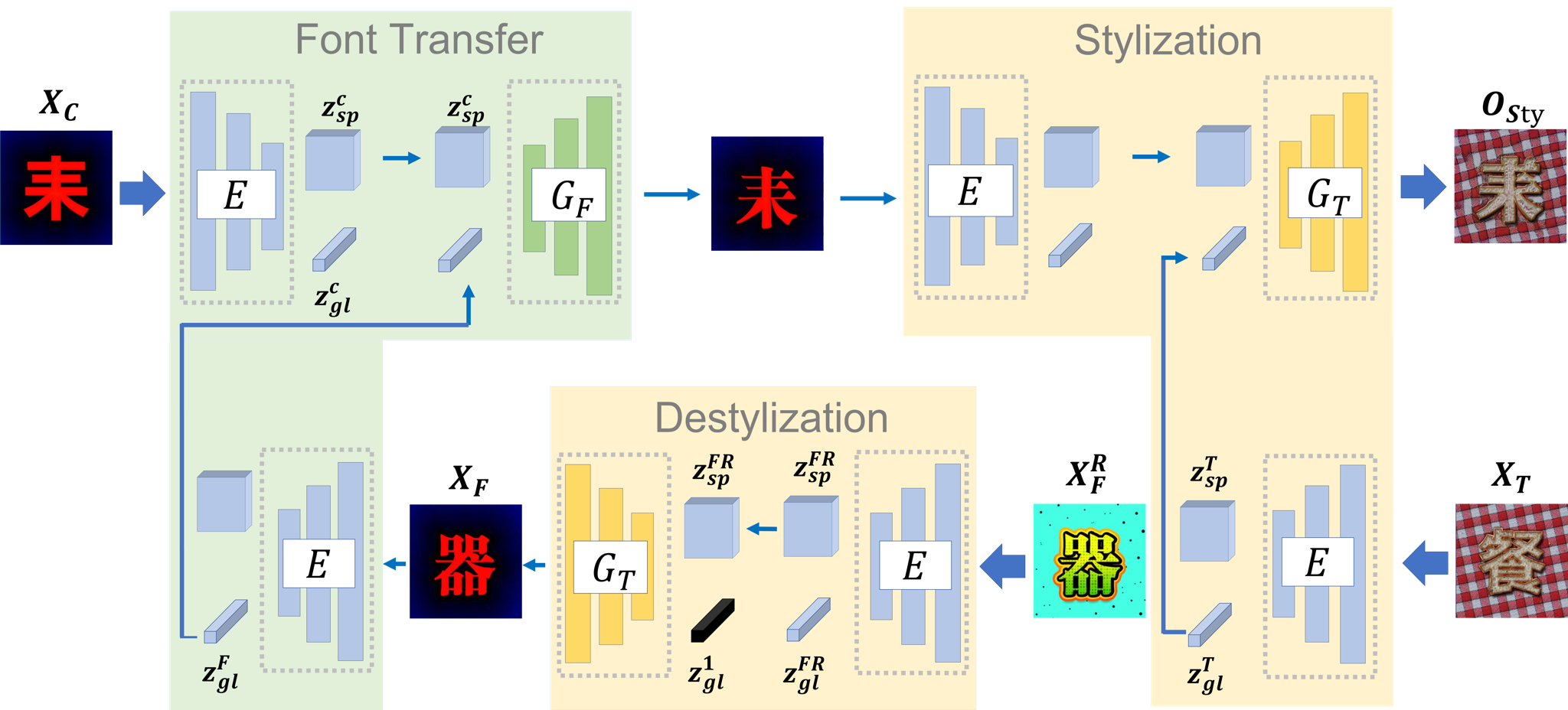}
\vspace{-10pt}
\caption{The overall architecture and inference pipeline of GenText. $X_F^R$, $X_C$, $X_T$ and $O_{Sty}$ denote the font reference image, content image, texture image and the output image of stylization stage, respectively. The $z_{sp}$ and $z_{gl}$ is the spatial code and global code while $z_{gl}^1$ is the constant vector filled by $1$. The “FR”, “T”, “C” {represent} “font reference”, “texture”, and 
“content”, respectively. }
\label{fig:inference}
\vspace{-15pt}
\end{figure*}

\section{Related Work}


\subsection{Image to Image Generation}
With the progress of neural network's powerful expression ability, \cite{gatys2016image} pioneered a neural algorithm of image style based on CNN that could separate the content and style of the image then applied them to new images. But this method required a slow calculation process and could not be applied to arbitrary new styles. To address these limitations, \cite{Huang_2017_ICCV} proposed a novel adaptive instance normalization (AdaIN) layer, aligned the mean and variance of content features with those of style features, for arbitrary style transformations in real-time. Moreover, driven by the development of GAN \cite{goodfellow2014generative,mirza2014conditional,isola2017image,zhu2017unpaired,Huang_2018_ECCV,Liu_2019_ICCV,Choi_2020_CVPR}, \cite{karras2017progressive} generated high-quality images through progressive network training. Benefiting from ProGAN, \cite{karras2019style} proposed StyleGAN, which decoupled an input latent vector in the hidden layer space to control various attributes and small stochastic variation. A subsequent work, StyleGAN2, was proposed by \cite{Karras_2020_CVPR}, which redefined generator normalization and added path regularization for better training process.

\subsection{Text Style Transfer}
Transferring artistic text font and texture while maintaining content information is extremely challenging. There are many efforts \cite{zhang2018separating,te141k,krishnan2021textstylebrush,zhan2019spatial,yang2019tet,yang2019controllable,yang2018context} to migrate text style from some given character information. Some work focused on text editing, for instance,  \cite{wu2019editing} trained an SRNet to edit text styles in natural images. \cite{zhan2019spatial} presents an innovative Spatial Fusion GAN (SF-GAN) that used two different generators to achieve image coupling in geometry and appearance spaces. \cite{roy2020stefann} proposed two architectures, FANnet and Colornet, to preserve source font and source color. Meanwhile, \cite{yang2020swaptext} presented the image-based SwapText to transfer texts across scene images. More recently, \cite{krishnan2021textstylebrush} introduced TextStyleBrush to transfer source style to new content by disentangling the content of a text image in a self-supervised manner.

Furthermore, other work focused on artistic text style transfer. Some efforts \cite{upchurch2016z,fogel2020scrabblegan,davis2020text,kang2020ganwriting,gomez2019selective,alonso2019adversarial} addressed font migration.
\cite{lyu2017auto} used GAN directed by the auto-encoder to generate Chinese calligraphy characters in a specified style. \cite{cha2020few} proposed Dual Memory-augmented Font Generation Network (DM-Font), which {employed} memory components and global-context awareness in the generator to generate a {high-quality} font library from only a few samples. In the meantime, inspired by low-rank matrix factorization, \cite{park2020few} designed a novel font generation method, which synthesized the local details of text by localized style representations and factorization. Other efforts addressed artistic text style transfer. For example,  \cite{yang2016awesome,yang2018context} proposed {a} patch-based texture synthesis model that matched stylized patches position to generator image blocks, which was affected by font structure differences and thus suffered from a heavy computational burden. Driven by the improvement of GAN and computing power, \cite{azadi2017multicontent} used MC-GAN to accomplish the few-shot font style transfer task by transferring English alphabet glyph and texture. Meanwhile, TET-GAN\cite{yang2019tet} was designed to use two subnetworks to support stylization and destylization for text effects transfer. Unlike the aforementioned methods, \cite{yang2019controllable} proposed a novel bidirectional shape matching framework with a scale-controllable module to establish {a} glyph-style transfer.

\begin{figure*}[t]
\centering
\includegraphics[scale=0.14]{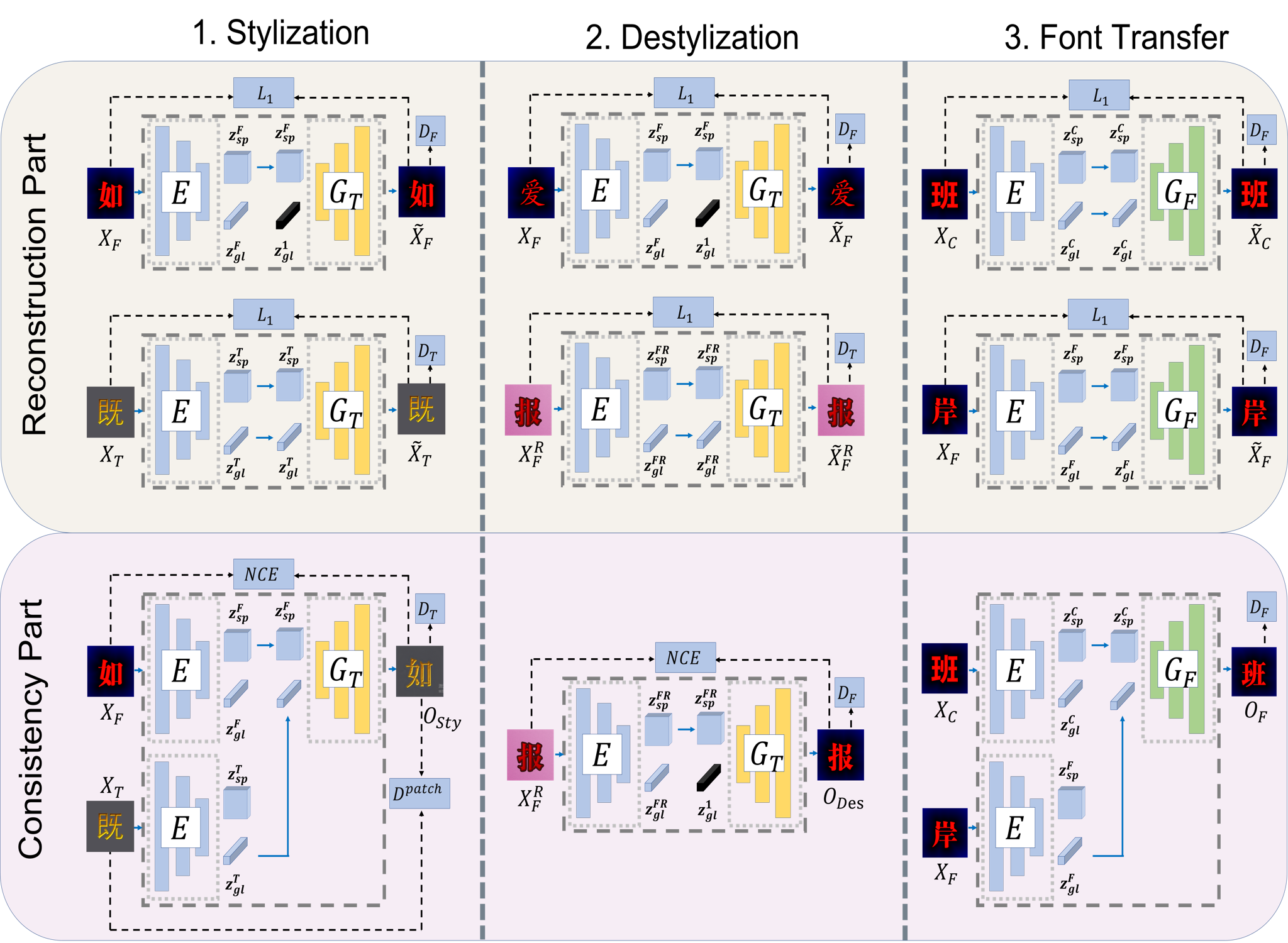}
\vspace{-10pt}
\caption{The training process in different stages.The detailed definition of each symbol is provided in Supplementary Material. 
}
\label{fig:train}
\vspace{-15pt}
\end{figure*} 


\section{Methodology}
In this section, we provide a detailed description about our unsupervised artistic text generation model. 
We provide a symbol list with the detailed definition in Supplementary Material to increase the {readability} of our paper. 
\subsection{Overall}
\label{sec:overall}
As shown in Fig. \ref{fig:overview}, to provide a more flexible adjustment for text generation, we select content, font, and texture as three separate components for artistic text. 
Since the font image can be generated by de-rendering the artistic text, in this paper, the font image $X_F$ refers to the font image shown in Fig. \ref{fig:overview} while the font reference image $X_F^R$ refers to the artistic text which {provides} the font information for our generation model. 
Based on the reference images, our model can {effectively} render the character specified by the content image in the corresponding font and texture styles. 

To achieve {the} above targets, our central contribution is to develop a unified model to simultaneously accomplish three sub-tasks: one to de-render the artistic style in font reference image for destylization, one to convert the font of the content image to be consistent with the font image for font transfer, and {another} to render the resulted font image in the style specified by the texture reference image for stylization. 
Fig. \ref{fig:inference} exhibits the overall architecture and inference pipeline of our proposed model, which consists of one powerful encoder $E$, two separate generators $\{G_{F}, G_{T}\}$, one for font transfer, the other for destylization and stylization. 
For each input reference image, the encoder $E$ outputs two latent codes, one is the spatial code $z_{sp}$ for embedding structure relations, and the other is the global code $z_{gl}$ for embedding style information, which can be expressed as $z_{sp}, z_{gl} = E(x)$. 
The training steps of stylization, destylization, and font transfer are separately introduced below. 

\subsection{Stylization}
\label{sec:Stylization}
As shown in Fig. \ref{fig:inference}, in this paper, the stylization is defined as rendering the font images $X_F$ in the style specified by a texture reference text $X_T$, where the structure information should remain the same as font image $X_F$ while the texture style should come from the reference image $X_T$. 
The $X_{F}$ and $X_{T}$ are passed through the encoder $E$ to obtain the latent code $z_{sp}^F, z_{gl}^F = E(X_F)$ and $z_{sp}^T, z_{gl}^T = E(X_T)$, respectively. 
Based on above representations, generator $G_{T}$ receives the latent code $z_{sp}^F$ and $z_{gl}^T$ to produce the stylized text ${O}_{Sty}$, which can be expressed as ${O}_{Sty} = G_{T}(z_{sp}^F, z_{gl}^T)$. 

Different from existing artistic text style transfer models, to efficiently utilize all artistic text images, the unsupervised learning strategy is implemented in our optimization process. 
In {the} optimization process, the central problem is to effectively learn a spatial code for encoding structure information and a global code for embedding style {characteristics}.

\textbf{Reconstruction Losses.} 
To make the complete information embedded in the spatial code and style code, we optimize our model via {the} reconstruction task. 
Specially, the font image $X_{F}$ and the texture reference image $X_{T}$ are reconstructed toward $\widetilde{X}_F = G_T(z_{sp}^T, z_{gl}^1)$ and $\widetilde{X}_T = G_T(z_{sp}^T, z_{gl}^T)$, respectively. 
Since the style generator $G_T$ is designed for transferring texture styles, to avoid the disturbance of various font styles, we replace $z_{gl}^T$ by $z_{gl}^1$ to reconstruct the font image, where $z_{gl}^1$ is a constant global vector and each element is filled by $1$. 
Then the $L_1$ loss is utilized to force the network to completely reconstruct input images. 
Since the $L_1$ loss tends to reconstruct low-frequency signals in the image, which leads to blurry images, we further employ two domain discriminators $D_{F}$ and $D_{T}$ to recover high-frequency information \cite{isola2017image} in $\widetilde{X}_F$ and $\widetilde{X}_{T}$, respectively. 
The reconstruction losses can be formulated as: 
\begin{small}
\begin{align}
\mathcal{L}_{Rec}^{Sty} =& \mathbb{E}_{X_{F}} \left[\|\widetilde{X}_F-{X}_F\|_{1} + D_{F}(\widetilde{X}_{F}) \right] \notag \\ 
 & + \mathbb{E}_{X_{T}} \left[ \|\widetilde{X}_T-{X}_T\|_{1} + D_{T}(\widetilde{X}_{T}) \right] .
\end{align}
\end{small}

\textbf{Consistency Losses.} 
The reconstruction loss optimizes spatial code and global code in a coherent manner. In this section, we introduce the consistency constraints to make the spatial code only focus on structure information while global code only focuses on style {characteristics}.


Specially, for spatial code, we implement structure consistency constraints on $O_{Sty}$ with $X_{F}$ by $NCE$ loss\cite{park2020contrastive}. 
The consistency of the overall structure between two images can be equivalent to the consistency of each local structure in the same position. 
Based on this motivation, for a query pixel in the feature map of $O_{Sty}$, we select the pixel at the same position in the feature map of $X_F$ as positive examples, while other pixels as negative examples. 
After the fully connected layer, we calculate the $NCE$ loss in the embedded space, which can be {expressed} as:
\begin{small}
\begin{align}
NCE(O_{Sty}, X_f) = &
- \log \left[\frac{e^{\left(\boldsymbol{v} \cdot \boldsymbol{v}^{+} / \tau\right)}}{e^{\left(\boldsymbol{v} \cdot \boldsymbol{v}^{+} / \tau\right)}+\sum_{n=1}^{N} e^{\left(\boldsymbol{v} \cdot \boldsymbol{v}_{n}^{-} / \tau\right)}}\right] ,
\end{align}
\end{small}
where $\boldsymbol{v}, \boldsymbol{v}^{+}$, and $\boldsymbol{v}^{-}_n$ are K-dimensional vectors generated from $O_{Sty}$ and $X_F$, which denote the features of query, positive, and $n$-th negative samples, respectively. 

For global code, we implement texture consistency constraints on $O_{Sty}$ with $X_{T}$ by $D^{patch}$ loss.
The purpose of the patch discriminator is to constrain texture consistency between $O_{Sty}$ and $X_T$ by making the patches derived from $O_{Sty}$ indistinguishable with the patches derived from $X_T$.


In addition, we also utilize the domain discriminator $D_T$ to constrain the domain consistency of $O_{Sty}$ with all texture images in the training set.

Based on {the} above discussions, the consistency losses can be expressed as: 
\begin{small}
\begin{align}
\mathcal{L}_{Cons}^{Sty} =& \mathbb{E}_{(X_F,X_T)} \left[NCE(O_{Sty}, X_{F}) + D_{T}(O_{Sty}) \right.  \notag \\
 & \left. + {D}^{patch}(O_{Sty}, X_{T}) \right] .
\end{align}
\end{small}





\subsection{Destylization}
\label{sec:Destylization}
In this paper, the destylization is defined as generating the font images from the reference artistic texts (also called font reference images in our terminology discussed in the first paragraph in Sect. \ref{sec:overall}) towards removing texture effect. 
To incorporate the destylization and stylization into a unified platform, we define the style of font image as a specific texture style, namely font-type texture, and utilize the constant global code $z_{gl}^1$ to keep consistent with the font reconstruction task in stylization stage. 
With this definition, we regard destylization as a special kind of stylization in this paper and utilize a similar training process as stylization. 
However, there is a key difference between destylization and stylization, which the target texture in stylization is provided by the texture reference images while the target texture in destylization is the pre-defined font-type style. 
This difference leads to some modifications in {the} training process, which will be discussed in the following.  

As shown in Fig. \ref{fig:train}, to better optimize the destylization stage, the font image $X_{F}$ and the font reference image $X_F^R$ are both utilized in {the} training process. 
Similar {to} stylization, apart from the resulted destylized image $O_{Des}$, we generate the reconstructed images $\widetilde{X}_{F}$ and $\widetilde{X}_{F}^R$ to optimize our model under the reconstruction losses:
\begin{small}
\begin{align}
\mathcal{L}_{Rec}^{Des}=&
\mathbb{E}_{X_{F}} \left[ \|\widetilde{X}_{F}-{X}_{F}\|_{1} + D_{F}(\widetilde{X}_{F}) \right] \notag \\ 
& + \mathbb{E}_{X_{F}^R} \left[ \|\widetilde{X}_{F}^R-{X}_{F}^R\|_{1} + D_{T}(\widetilde{X}_{F}^R) \right] .
\end{align}
\end{small}



\textbf{Consistency Losses.} We utilize the $NCE$ loss to maintain structure consistency in detylization task since the detylization task is required by removing texture style while maintaining the structure of ${X}_{F}^R$. 
Moreover, there is only one target texture, namely the font-type texture, thus only the domain discriminator $D_F$ is needed for texture consistency. 
Based on {the} above discussions, the consistency losses in detylization task can be expressed as
\begin{small}
\begin{equation}
\setlength{\belowdisplayskip}{3pt}
\mathcal{L}_{Cons}^{Des}=\mathbb{E}_{(X_F,X_F^R)} [NCE(O_{Des}, X_{F}^R) + D_{F}(O_{Des}) ] .
\end{equation}
\end{small}










\subsection{Font Transfer}
\label{sec:Font Transfer}
In this paper, we consider the font transfer task as rendering defined by the content image $X_C$ defined by a standard font style in the style specified by the font images $X_F$, which can also be regarded as a special stylization task by rendering current content image with the font style according to the reference image. 
We utilize the same encoder $E$ to project the font style code into the same feature space with the texture style code, and then employ two different generators to decode font and texture style into different domains. 

As shown in Fig. \ref{fig:train}, we employ the font image $X_{F}$ and the content image $X_C$ defined by a standard font style to optimize the font transfer stage. 
The training data $X_{F}$ and $X_{C}$ are passed through the encoder $E$ to obtain the latent code $z_{sp}^F, z_{gl}^F = E(X_F)$ and $z_{sp}^C, z_{gl}^C = E(X_C)$, respectively. 
Based on above representations, the font generator $G_{F}$ can generate the reconstructed font image $\widetilde{X}_{F}$, content image $\widetilde{X}_{C}$ and the font-transferred image $O_{F}$ via the processes $\widetilde{X}_{F} = G_F(z_{sp}^F, z_{gl}^F)$, $\widetilde{X}_{C} = G_F(z_{sp}^C, z_{gl}^C)$ and $O_{F} = G_F(z_{sp}^C, z_{gl}^F)$, respectively. 


Similar {to} other stages, we also implement the reconstruction loss and font discriminator loss to optimize our model, which can be expressed as: 
\begin{small}
\begin{align}
\setlength{\belowdisplayskip}{3pt}
\mathcal{L}_{Rec}^F=&
\mathbb{E}_{X_{C}} \left[ \|\widetilde{X}_C-X_{C}\|_{1} + D_{F}(\widetilde{X}_C) \right] \notag \\ 
& + \mathbb{E}_{X_{F}} \left[ \|\widetilde{X}_F-X_{F}\|_{1} + D_{F}(\widetilde{X}_F) \right] .
\end{align}
\end{small}

There are two essential differences between font transfer and stylization tasks. 
First of all, in terms of structural information, stylization needs to maintain structural consistency with the input image, while the essence of font transfer is to make a certain degree of deformation related to the characteristics of the special font. 
Secondly, the difference between different fonts mainly comes from large-scale {strokes}, where the local patches are not efficient to discriminate them. 
Therefore, applying $NCE$ loss on $O_F$ is harmful while $D^{patch}$ loss is not necessary. 
We will verify the above statements in ablation study.

In addition, the domain discriminator $D_{F}$ should be employed to perform domain constraint on the output image $O_F$, which leads to the consistency loss in font transfer task as
\begin{small}
\begin{equation}
\mathcal{L}_{Cons}^F=\mathbb{E}_{(X_C,X_F)}[D_{F}(O_{F})]. 
\end{equation}
\end{small}
\subsection{End-to-End Optimization and Inference}
\label{sec3.5}
Our model provides a flexible solution for artistic text generation task. 
On one hand, people can optimize and utilize each stage separately to finish each individual task, whose training and inference processes have been detailed discussed in the previous section. 
On the other hand, our model can be {jointly} optimized and utilized as a uniform model. 
The overall training process of the end-to-end model can be expressed as:
\begin{small}
\begin{equation}
\setlength{\belowdisplayskip}{3pt}
\setlength{\belowdisplayskip}{3pt}
\mathcal{L} = \mathcal{L}_{Rec}^{Sty} + \mathcal{L}_{Rec}^{Des} + \mathcal{L}_{Rec}^{F} + \mathcal{L}_{Cons}^{Sty} + \mathcal{L}_{Cons}^{Des} + \mathcal{L}_{Cons}^{F} .
\end{equation}
\end{small}


Based on the above joint optimization, our model can efficiently perform flexible adjustment in artistic text generation task. 
Fig. \ref{fig:inference} shows the inference pipeline for this process. 
Firstly, a font reference image $X_F^R$ is passed to the destylization stage to generate the font image. 
Then the resulted font image together with the content reference image $X_C$ are further passed to the font transfer stage to render the target content with the font style. 
Finally, the stylization stage is employed to render the resulted image in texture style specified by the texture reference image $X_T$. 
According to the above steps, one can flexible adjust the font and texture styles in arbitrary text, which is a major contribution of our current work.

\section{Experiments}
In this section, we perform qualitative and quantitative study to validate the effectiveness of our proposed artistic text generation model. 
\subsection{Dataset}
TE141K \cite{te141k} is the largest artistic text database with $152$ different artistic textures, including $65$ Chinese textures, $67$ English textures, and $20$ special symbol textures. 
This database provides the corresponding font images for each texture image, but such paired supervision is not utilized in our model. 
We select all font images and texture images in TE141K to form font domain data $X_{F}$ and texture domain data $X_{T}$, respectively. 
Moreover, the Microsoft Yahei font is utilized as the content domain data $X_C$ to provide the target character information in our model. 

\begin{figure}[t]
\centering
\includegraphics[scale=0.25]{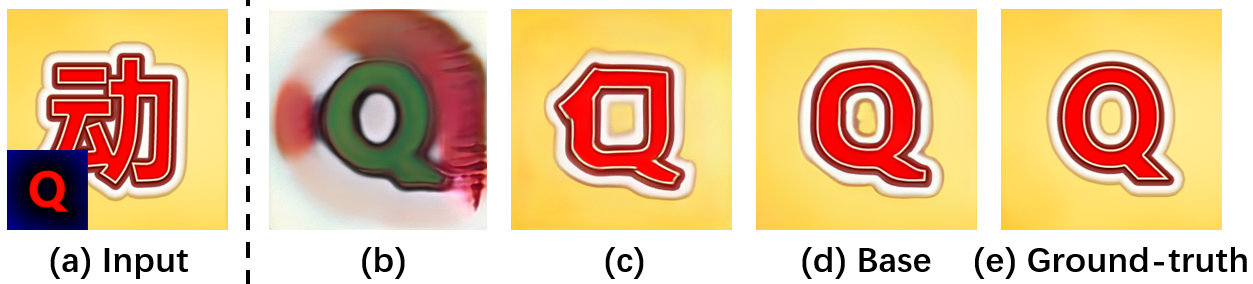}
\vspace{-10pt}
\caption{Effect of $NCE$ loss and $D^{patch}$ loss on stylization task. (a) Input. (b) w/o $D^{patch}$ loss. (c) w/o $NCE$ loss. (d) Base model. (e) Ground truth.}
\label{fig:stylization_abalation_study}
\vspace{-18pt}
\end{figure} 

To compare with existing artistic text effect transfer models introduced in \cite{te141k}, the independent stylization stage in our paper is trained on all TE141K data, that is, including Chinese characters, English alphabets, and special symbols. 
However, since the content domain data $X_C$ only contains Chinese characters, for the font transfer model and the joint-optimized model, we only use Chinese data to perform the ablation study. 


\subsection{Implementation Details}
Due to the page limitation, we offer a brief introduction of our network architecture and the optimization settings.
We implement our code based on Ref. \cite{swapping}.
The detailed description is provided in Supplementary Material. 

\textbf{Network Architecture.} 
The backbone of the encoder network $E$ includes four downsampling residual blocks, followed by two convolution layers. 
Then the spatial code and global code are generated from two separate branches. 
The generator $G_{T}$ and $G_{F}$ have the same network structure with different weights, and each of them consists of two residual blocks to maintain resolution and four upsampling residual blocks. 
The font discriminator $D_{F}$ and texture discriminator $D_{T}$ also have the same network structure with different weights, which is identical to StyleGAN2 \cite{Karras_2020_CVPR}. 

\textbf{Network Training.}
All training images are resized to $256\times 256$.
We utilize ADAM\cite{kingma2014adam} to optimize our model with the parameters $lr=0.002$, $\beta_{1}=0.0$ and $\beta_{2}=0.99$. 
The batch size is set as $bs=16$ in all experiments, which can be optimized on 8 GeForce RTX 2080 Ti GPUs.
We use non-saturating GAN loss\cite{goodfellow2014generative} and lazy $R_1$ regularization\cite{mescheder2018training,karras2020analyzing} in all discriminators, including domain discriminators and patch discriminators. 

\subsection{Evaluation Metrics} 
In this paper, to quantitative evaluate our proposed method, four commonly used measurements \cite{krishnan2021textstylebrush,te141k} are employed, including (1) Peak Signal-to-Noise Ratio (PSNR), (2) Structural Similarity Index Measure (SSIM), (3) perceptual loss, and (4) Style loss. 
Among them, the higher PSNR and SSIM scores indicate better image quality, while the lower perceptual loss and style loss indicate better semantic similarity and style similarity of two images, respectively. 




\begin{table}[htbp]
	\centering
	\vspace{-10pt}
	\begin{small}
	\caption{Ablation study on the stylization part of GenText with the measurements of PSNR, SSIM, Perceptual loss, Style loss.}
	\vspace{-10pt}
	\label{tab:Abalation4Sty}  
	\setlength{\tabcolsep}{0.75mm}{
	\begin{tabular}{ccccccc}
		\hline\noalign{\smallskip}	
		Model & PSNR$\uparrow$ & SSIM$\uparrow$ & Perceptual$\downarrow$ & Style$\downarrow$  \\
		\noalign{\smallskip}\hline\noalign{\smallskip}
		Base & \textbf{19.415} & \textbf{0.781} & \textbf{1.141} & \textbf{0.0014}   \\
		w/o $NCE$ loss & 18.826 & 0.769 & 1.234 & 0.0016 \\
		w/o $D^{patch}$ loss & 10.013 & 0.429 & 1.683 & 0.0044 \\
		\noalign{\smallskip}\hline
	\end{tabular}}
	\end{small}
	\vspace{-15pt}
\end{table}

\subsection{Stylization Task}
Due to the widely potential application of stylization and font transfer, we perform the ablation study to investigate the stylization task and font transfer task in this paper. 
In this section, we focus on the stylization task and optimize the stylization part of our GenText to render the font images in the style of the texture reference image. 
\subsubsection{Ablation Study on Stylization Task}
In this subsection, we select the stylization part of our GenText (introduced in Sect. \ref{sec:Stylization}) as our base model and perform the ablation study with different settings. 

We first focus on the consistency constraints on the stylization task. 
Since we implement $NCE$ loss to maintain structural consistency and $D^{patch}$ loss to maintain texture consistency, we separately remove the above objective functions to investigate the influence of consistency constraints in our model. 
The experimental and visualization results are shown in Tab. \ref{tab:Abalation4Sty} and Fig. \ref{fig:stylization_abalation_study}, respectively. 
We can draw the following conclusions: 
(1). Without structural consistency (marked as w/o $NCE$ loss), the structure information cannot be effectively transferred, and the PSNR has a significant drop due to the mismatch of structures between the generated image and ground truth image. 
For example, when the texture of Chinese characters is moved to English characters, the local bending structure of English characters will be destroyed without $NCE$ loss. 
(2). Without texture consistency (marked as w/o $D^{patch}$ loss), the effective representation of the texture cannot be learned, resulting in a comprehensive decline in the measures.

\begin{table}[htbp]
    \vspace{-10pt}
	\centering
	\begin{small}
	\caption{Performance on the stylization part of GenText  with the measurement of PSNR, SSIM, Perceptual loss, Style loss, respectively. The upper part is the supervised models while the lower part is the unsupervised models.}
	\vspace{-10pt}
	\label{tab:SOTA}  
	\setlength{\tabcolsep}{0.75mm}{
	\begin{tabular}{ccccccc}
		\hline\noalign{\smallskip}	
		Model & PSNR$\uparrow$ & SSIM$\uparrow$ & Perceptual$\downarrow$ & Style$\downarrow$  \\
		\noalign{\smallskip}\hline\noalign{\smallskip}
		Doodles\cite{champandard2016semantic} & 18.172 & 0.666 & 1.5763 & 0.0031 \\
		T-Effect\cite{yang2016awesome} & 21.402 & 0.793 & 1.0918 & 0.0020 \\
		Pix2pix\cite{isola2017image} & 20.518 & 0.798 & 1.3940 & 0.0032 \\
		BicycleGAN\cite{zhu2017multimodal} & 20.950 & 0.803 & 1.5080 & 0.0033 \\
		TET-GAN\cite{yang2019tet} & \textbf{27.626} & \textbf{0.900} & 0.8250 & 0.0014 \\
		Ours & 24.645 & 0.874 & \textbf{0.7552} & \textbf{0.0009} \\
		\noalign{\smallskip}\hline\noalign{\smallskip}
		AdaIN\cite{Huang_2017_ICCV} & 13.939 & 0.612 & 1.6147 & 0.0038   \\
		WCT\cite{li2017universal} & 14.802 & 0.619 & 1.8626 & 0.0036 \\
		StarGAN\cite{Choi_2020_CVPR} & 14.977  & 0.612 & 1.9144 & 0.0045 \\
		Ours  & \textbf{19.415} & \textbf{0.781} & \textbf{1.1407} & \textbf{0.0014} \\
		\noalign{\smallskip}\hline
	\end{tabular}}
	\end{small}
	\vspace{-12pt}
	
\end{table}

\subsubsection{Comparison with State-of-the-Art.}
Since the stylization task in this paper is the text effect transfer defined in Ref. \cite{te141k}, we compare the stylization part with existing SOTA models in Tab. \ref{tab:SOTA}.
Compared with the unsupervised learning models, our method achieves the best results in all four measures. 
In addition, in terms of style loss, our model outperforms the current best supervised model, which proves that the texture style can be effectively transferred without any paired data. 

\begin{figure}[t]
\centering
\includegraphics[scale=0.21]{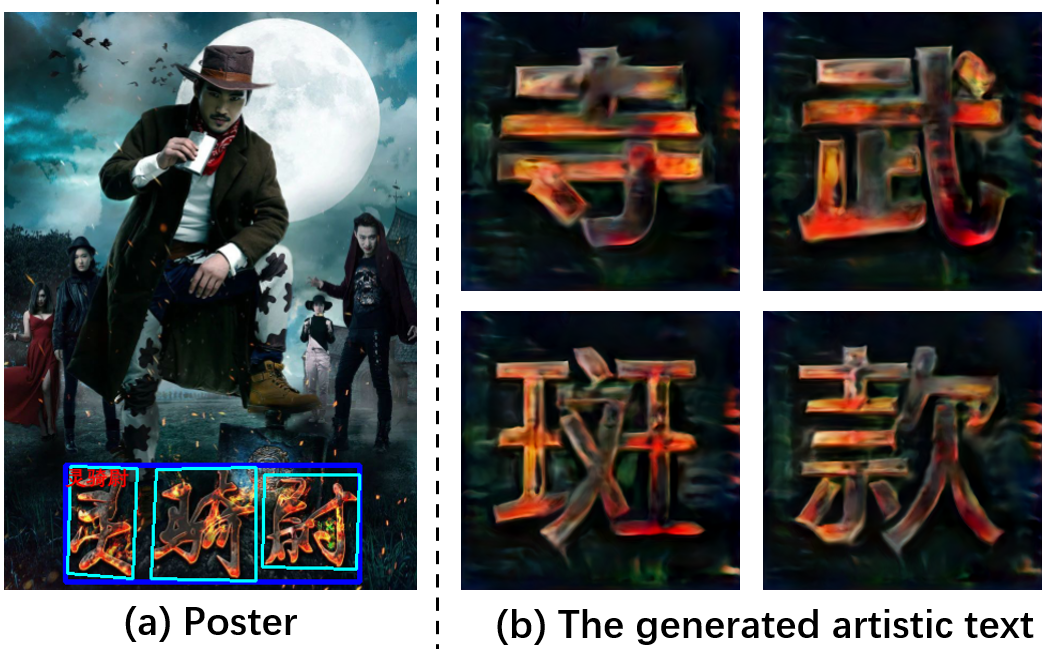}
\vspace{-10pt}
\caption{Visualization results for unsupervised finetune in the open scene. (a): The poster image collected from the Internet. (b): The generated artistic text by model finetuned on open scene data.}
\label{fig:Unsupervised finetune on open scene}
\vspace{-15pt}
\end{figure} 

\subsubsection{Generalization Ability}
To verify the generalization ability provided by the unsupervised optimization strategy, we test our stylization model in the open scene by rendering font images in the artistic style collected from some poster images. 
Specifically, $10$ different poster images are collected from the Internet, and each artistic character in the image is labeled with a quadrilateral.
Then the character image is cropped according to its corresponding minimum enclosing rectangle.
We employ the base model as the pre-trained model and finetune this model on the above artistic character images. 
Finally, we generate a set of artistic text images from the resulted model. 
Some stylized results are shown in Fig. \ref{fig:Unsupervised finetune on open scene}, which verify the effectiveness of our model. 


\subsubsection{Application}
Our stylization model can provide two additional useful applications, text interpolation, and text instance style editing. 

\begin{figure}[t]
\centering
\includegraphics[scale=0.23]{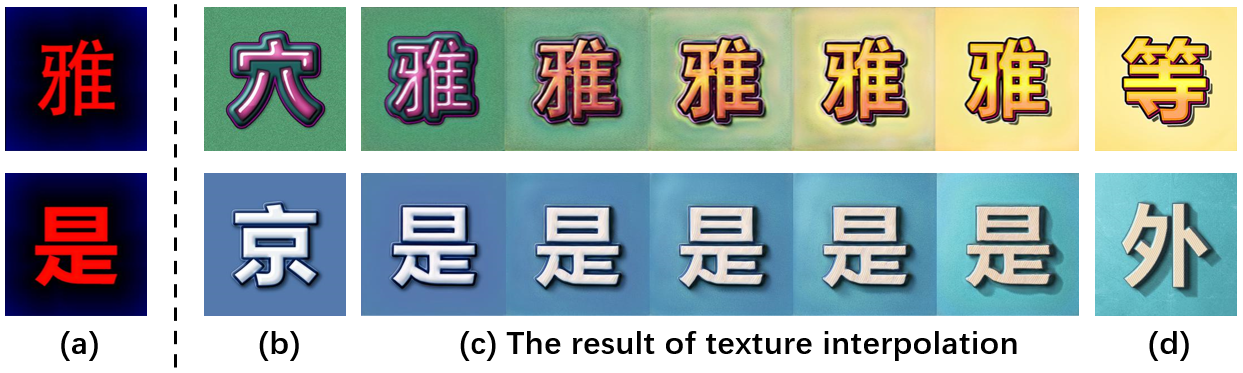}
\vspace{-10pt}
\caption{Visualization results for texture interpolation task. (a): input font image. (b): Input texture reference image A. (c): The generated results of texture interpolation between A and B. (d): Input texture reference image B. }
\label{fig:Texture interpolation.}
\vspace{-5pt}
\end{figure}

\textbf{Texture Interpolation. }
In our model, the texture style is represented by the global code $z_{gl}$, which is a vector without spatial dimension. 
Given one font image and two different texture reference images A and B, we can linearly interpolate their global code to obtain a series of global codes, which will generate a set of blend styles by blending two original textures to various degrees based on the different interpolation coefficients. 
As shown in Fig. \ref{fig:Texture interpolation.}, the flexible blended artistic text image can be generated from GenText.  

\begin{figure}[t]
\centering
\includegraphics[scale=0.28]{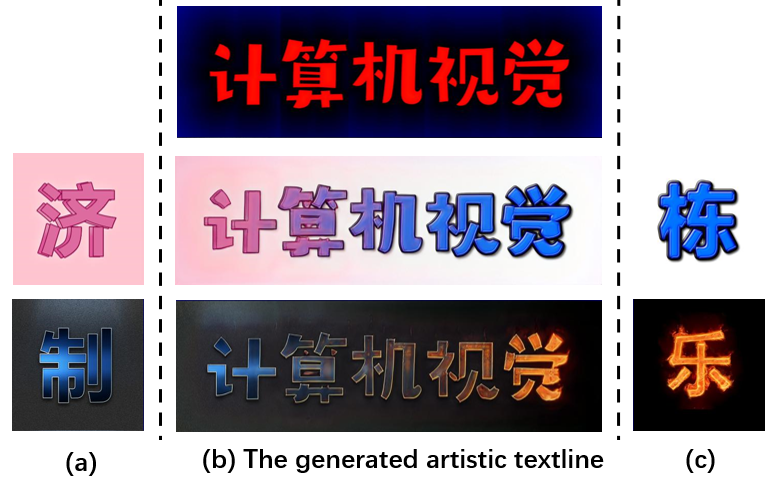}
\vspace{-10pt}
\caption{Visualization results for text instance style editing. The first row is an input textline image generated with cartoon font. (a): Left texture reference image. (b): The generated artistic textline with spatial gradient style. (c): Right texture reference image.}
\label{fig:texture spatial mix}
\vspace{-15pt}
\end{figure} 

\textbf{Text Instance Style Editing.}
Since our encoder and generator are fully convolutional structures, the text instance with multiple characters can be rendered by our model. 
As an application, we first obtain the spatial code $z_{sp}$ of the input text instances from the encoder $E$. 
Then we select two global codes $z_{gl}^{left}$ and $z_{gl}^{right}$ to gradual blend two texture styles. 
As shown in Fig. \ref{fig:texture spatial mix}, we employ the synthesized text instances with two different texture images to edit text instances with spatial gradient style. 

\subsection{Font Transfer Task}
\begin{figure}[t]
\centering
\includegraphics[scale=0.21]{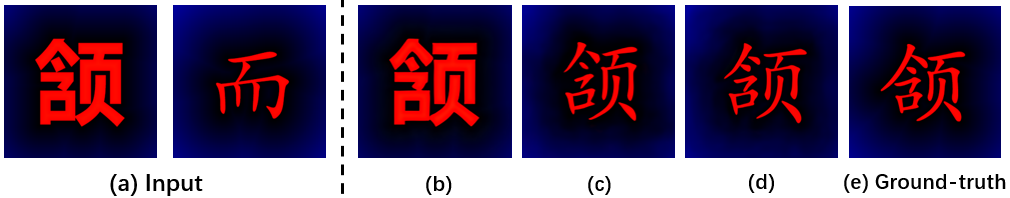}
\vspace{-10pt}
\caption{Effect of $NCE$ loss and $D^{patch}$ loss on font transfer task. (a): Input (left: content image, right: font image). (b): with $NCE$ loss. (c): with $D^{patch}$ loss. (d): Base model. (e): Ground truth.}
\label{fig:font transfer ablation study}
\vspace{-5pt}
\end{figure}

The font transfer task is another promising visual creation task, and we perform the ablation study on this task in this subsection. 
As discussed in Sect. \ref{sec:Font Transfer}, the $NCE$ loss and $D^{patch}$ loss are not suitable for the font transfer task. 
To verify the above statement, we separately implement above objective functions to investigate the influence of consistency constraints in our model. 
The base model is the font transfer part of our GenText. 
As shown in Tab. \ref{tab:ablation4font} and Fig. \ref{fig:font transfer ablation study}, we can draw the following conclusions: 
(1). The structural consistency (marked as + $NCE$ loss) makes the resulted image almost keep the same font style as the original content image, which causes the failure of the font transfer task. 
(2). The additional patch discriminator (marked as + $D^{patch}$ loss) does not work. 
We think the reason mainly comes from that the patch discriminator $D^{patch}$ cannot well discriminate different fonts from local patches. 
(3). The visualization results show the reconstruction loss can effectively optimize the font transfer task. 

\begin{table}[htbp]
	\centering
	\vspace{-3pt}
	\begin{small}
	\caption{Ablation study on the font transfer part of GenText with the measurements of PSNR, SSIM, Perceptual loss, Style loss.}
	\label{tab:ablation4font}  
	\vspace{-7pt}
	\setlength{\tabcolsep}{0.75mm}{
	\begin{tabular}{ccccccc}
		\hline\noalign{\smallskip}	
		Model & PSNR$\uparrow$ & SSIM$\uparrow$ & Perceptual$\downarrow$ & Style$\downarrow$  \\
		\noalign{\smallskip}\hline\noalign{\smallskip}
		Base & \textbf{21.427} & 0.946 & \textbf{0.6245} & \textbf{0.0007}   \\
		{+ $NCE$ loss} & 19.247 & 0.904 & 0.826 & 0.0012 \\
		{+ $D^{patch}$ loss} & 21.346 & \textbf{0.948} & 0.6327 & \textbf{0.0007} \\
		\noalign{\smallskip}\hline
	\end{tabular}}
	\end{small}
	\vspace{-15pt}
\end{table}

\subsection{End-to-End Model}

\begin{figure}[htbp]
\vspace{-15pt}
\centering
\includegraphics[scale=0.25]{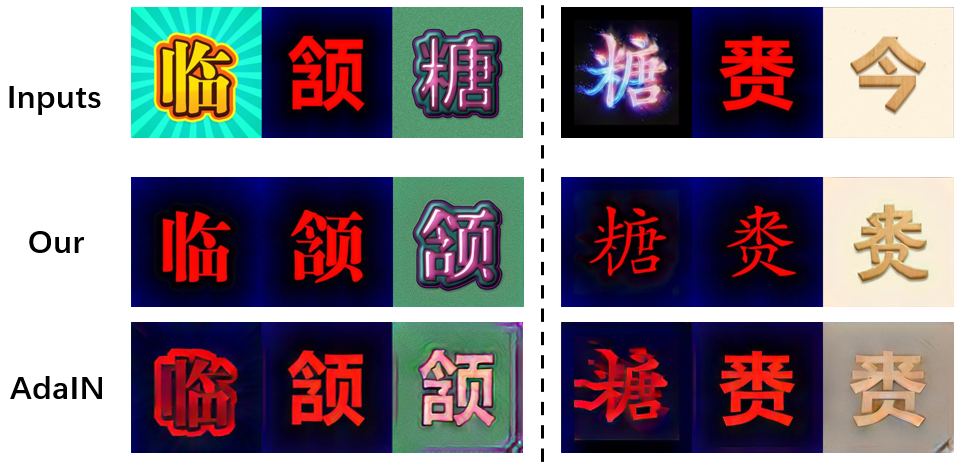}
\vspace{-10pt}
\caption{Visualization results for the End-to-End model. The first row is three input images, from left to right are font, content, and texture reference images. The second and third rows are the output images of our model and AdaIN model, respectively. The output images are arranged in the order of inference, which are the outputs of destylization, stylization, and font transfer tasks. }
\label{fig:end to end exp}
\vspace{-20pt}
\end{figure}

In this section, we employ our end-to-end model to perform flexible adjustment by generating an artistic character defined by content images with the font and texture styles specified by different reference images. 
To verify the effectiveness of our end-to-end model, we select AdaIN \cite{Huang_2017_ICCV} as the baseline for comparison. 
The visualization results are presented in Fig. \ref{fig:end to end exp}, which demonstrates that GenText achieve promising results than the baseline model and can effectively generate high-quality artistic text by rendering content images with different font and texture styles.
\section{Conclusion}
\setlength{\abovecaptionskip}{0.1cm}
\setlength{\belowcaptionskip}{-0.05cm}

In this paper, to solve the large diversity of artistic text and the difficulty of paired-data acquisition in artistic text generation task, we propose a flexible unsupervised artistic text generation model, namely GenText, to achieve controllable general artistic text generation based on the content, font and texture in reference text. 
Our GenText incorporate three different artistic editing tasks into a unified platform, where a powerful encoder network project the useful information into the spatial and style representations while two different generator are employed to generate artistic text via decoding global style code into font or texture domains. 
Qualitative and quantitative results demonstrate the superior performance of our proposed method.

\bibliographystyle{splncs04}
\bibliography{egbib}

\begin{thebibliography}{10}
\providecommand{\url}[1]{\texttt{#1}}
\providecommand{\urlprefix}{URL }
\providecommand{\doi}[1]{https://doi.org/#1}

\bibitem{alonso2019adversarial}
Alonso, E., Moysset, B., Messina, R.: Adversarial generation of handwritten
  text images conditioned on sequences. In: 2019 International Conference on
  Document Analysis and Recognition (ICDAR). pp. 481--486. IEEE (2019)

\bibitem{azadi2017multicontent}
Azadi, S., Fisher, M., Kim, V.G., Wang, Z., Shechtman, E., Darrell, T.:
  Multi-content gan for few-shot font style transfer. In: Proceedings of the
  IEEE conference on computer vision and pattern recognition. pp. 7564--7573
  (2018)

\bibitem{cha2020few}
Cha, J., Chun, S., Lee, G., Lee, B., Kim, S., Lee, H.: Few-shot compositional
  font generation with dual memory. In: Computer Vision--ECCV 2020: 16th
  European Conference, Glasgow, UK, August 23--28, 2020, Proceedings, Part XIX
  16. pp. 735--751. Springer (2020)

\bibitem{champandard2016semantic}
Champandard, A.J.: Semantic style transfer and turning two-bit doodles into
  fine artworks. arXiv preprint arXiv:1603.01768  (2016)

\bibitem{Choi_2020_CVPR}
Choi, Y., Uh, Y., Yoo, J., Ha, J.W.: Stargan v2: Diverse image synthesis for
  multiple domains. In: Proceedings of the IEEE/CVF Conference on Computer
  Vision and Pattern Recognition (CVPR) (June 2020)

\bibitem{davis2020text}
Davis, B., Tensmeyer, C., Price, B., Wigington, C., Morse, B., Jain, R.: Text
  and style conditioned gan for generation of offline handwriting lines. arXiv
  preprint arXiv:2009.00678  (2020)

\bibitem{fogel2020scrabblegan}
Fogel, S., Averbuch-Elor, H., Cohen, S., Mazor, S., Litman, R.: Scrabblegan:
  Semi-supervised varying length handwritten text generation. In: Proceedings
  of the IEEE/CVF Conference on Computer Vision and Pattern Recognition. pp.
  4324--4333 (2020)

\bibitem{gatys2016image}
Gatys, L.A., Ecker, A.S., Bethge, M.: Image style transfer using convolutional
  neural networks. In: Proceedings of the IEEE conference on computer vision
  and pattern recognition. pp. 2414--2423 (2016)

\bibitem{gomez2019selective}
Gomez, R., Biten, A.F., Gomez, L., Gibert, J., Karatzas, D., Rusi{\~n}ol, M.:
  Selective style transfer for text. In: 2019 International Conference on
  Document Analysis and Recognition (ICDAR). pp. 805--812. IEEE (2019)

\bibitem{goodfellow2014generative}
Goodfellow, I., Pouget-Abadie, J., Mirza, M., Xu, B., Warde-Farley, D., Ozair,
  S., Courville, A., Bengio, Y.: Generative adversarial nets. Advances in
  neural information processing systems  \textbf{27} (2014)

\bibitem{Huang_2017_ICCV}
Huang, X., Belongie, S.: Arbitrary style transfer in real-time with adaptive
  instance normalization. In: Proceedings of the IEEE International Conference
  on Computer Vision (ICCV) (Oct 2017)

\bibitem{Huang_2018_ECCV}
Huang, X., Liu, M.Y., Belongie, S., Kautz, J.: Multimodal unsupervised
  image-to-image translation. In: Proceedings of the European Conference on
  Computer Vision (ECCV) (September 2018)

\bibitem{isola2017image}
Isola, P., Zhu, J.Y., Zhou, T., Efros, A.A.: Image-to-image translation with
  conditional adversarial networks. In: Proceedings of the IEEE conference on
  computer vision and pattern recognition. pp. 1125--1134 (2017)

\bibitem{kang2020ganwriting}
Kang, L., Riba, P., Wang, Y., Rusi{\~n}ol, M., Forn{\'e}s, A., Villegas, M.:
  Ganwriting: Content-conditioned generation of styled handwritten word images.
  In: European Conference on Computer Vision. pp. 273--289. Springer (2020)

\bibitem{karras2017progressive}
Karras, T., Aila, T., Laine, S., Lehtinen, J.: Progressive growing of gans for
  improved quality, stability, and variation. arXiv preprint arXiv:1710.10196
  (2017)

\bibitem{karras2019style}
Karras, T., Laine, S., Aila, T.: A style-based generator architecture for
  generative adversarial networks. In: Proceedings of the IEEE/CVF Conference
  on Computer Vision and Pattern Recognition. pp. 4401--4410 (2019)

\bibitem{Karras_2020_CVPR}
Karras, T., Laine, S., Aittala, M., Hellsten, J., Lehtinen, J., Aila, T.:
  Analyzing and improving the image quality of stylegan. In: Proceedings of the
  IEEE/CVF Conference on Computer Vision and Pattern Recognition (CVPR) (June
  2020)

\bibitem{karras2020analyzing}
Karras, T., Laine, S., Aittala, M., Hellsten, J., Lehtinen, J., Aila, T.:
  Analyzing and improving the image quality of stylegan. In: Proceedings of the
  IEEE/CVF Conference on Computer Vision and Pattern Recognition. pp.
  8110--8119 (2020)

\bibitem{kingma2014adam}
Kingma, D.P., Ba, J.: Adam: A method for stochastic optimization. arXiv
  preprint arXiv:1412.6980  (2014)

\bibitem{krishnan2021textstylebrush}
Krishnan, P., Kovvuri, R., Pang, G., Vassilev, B., Hassner, T.: Textstylebrush:
  Transfer of text aesthetics from a single example. arXiv preprint
  arXiv:2106.08385  (2021)

\bibitem{li2017universal}
Li, Y., Fang, C., Yang, J., Wang, Z., Lu, X., Yang, M.H.: Universal style
  transfer via feature transforms. arXiv preprint arXiv:1705.08086  (2017)

\bibitem{Liu_2019_ICCV}
Liu, M.Y., Huang, X., Mallya, A., Karras, T., Aila, T., Lehtinen, J., Kautz,
  J.: Few-shot unsupervised image-to-image translation. In: Proceedings of the
  IEEE/CVF International Conference on Computer Vision (ICCV) (October 2019)

\bibitem{lyu2017auto}
Lyu, P., Bai, X., Yao, C., Zhu, Z., Huang, T., Liu, W.: Auto-encoder guided gan
  for chinese calligraphy synthesis. In: 2017 14th IAPR International
  Conference on Document Analysis and Recognition (ICDAR). vol.~1, pp.
  1095--1100. IEEE (2017)

\bibitem{mescheder2018training}
Mescheder, L., Geiger, A., Nowozin, S.: Which training methods for gans do
  actually converge? In: International conference on machine learning. pp.
  3481--3490. PMLR (2018)

\bibitem{mirza2014conditional}
Mirza, M., Osindero, S.: Conditional generative adversarial nets. arXiv
  preprint arXiv:1411.1784  (2014)

\bibitem{park2020few}
Park, S., Chun, S., Cha, J., Lee, B., Shim, H.: Few-shot font generation with
  localized style representations and factorization. arXiv preprint
  arxiv:2009.11042  (2020)

\bibitem{park2020contrastive}
Park, T., Efros, A.A., Zhang, R., Zhu, J.Y.: Contrastive learning for unpaired
  image-to-image translation. In: European Conference on Computer Vision. pp.
  319--345. Springer (2020)

\bibitem{swapping}
Park, T., Zhu, J.Y., Wang, O., Lu, J., Shechtman, E., Efros, A.A., Zhang, R.:
  Swapping autoencoder for deep image manipulation. In: Advances in Neural
  Information Processing Systems (2020)

\bibitem{roy2020stefann}
Roy, P., Bhattacharya, S., Ghosh, S., Pal, U.: Stefann: scene text editor using
  font adaptive neural network. In: Proceedings of the IEEE/CVF Conference on
  Computer Vision and Pattern Recognition. pp. 13228--13237 (2020)

\bibitem{upchurch2016z}
Upchurch, P., Snavely, N., Bala, K.: From a to z: supervised transfer of style
  and content using deep neural network generators. arXiv preprint
  arXiv:1603.02003  (2016)

\bibitem{wu2019editing}
Wu, L., Zhang, C., Liu, J., Han, J., Liu, J., Ding, E., Bai, X.: Editing text
  in the wild. In: Proceedings of the 27th ACM international conference on
  multimedia. pp. 1500--1508 (2019)

\bibitem{dgfont}
Xie, Y., Chen, X., Sun, L., Lu, Y.: Dg-font: Deformable generative networks for
  unsupervised font generation. CoRR  \textbf{abs/2104.03064} (2021),
  \url{https://arxiv.org/abs/2104.03064}

\bibitem{yang2020swaptext}
Yang, Q., Huang, J., Lin, W.: Swaptext: Image based texts transfer in scenes.
  In: Proceedings of the IEEE/CVF Conference on Computer Vision and Pattern
  Recognition. pp. 14700--14709 (2020)

\bibitem{yang2016awesome}
Yang, S., Liu, J., Lian, Z., Guo, Z.: Awesome typography: Statistics-based text
  effects transfer. In: Proceedings of the IEEE Conference on Computer Vision
  and Pattern Recognition. pp. 7464--7473 (2017)

\bibitem{yang2019tet}
Yang, S., Liu, J., Wang, W., Guo, Z.: Tet-gan: Text effects transfer via
  stylization and destylization. In: Proceedings of the AAAI Conference on
  Artificial Intelligence. vol.~33, pp. 1238--1245 (2019)

\bibitem{yang2018context}
Yang, S., Liu, J., Yang, W., Guo, Z.: Context-aware text-based binary image
  stylization and synthesis. IEEE Transactions on Image Processing
  \textbf{28}(2),  952--964 (2018)

\bibitem{te141k}
Yang, S., Wang, W., Liu, J.: Te141k: artistic text benchmark for text effect
  transfer. IEEE transactions on pattern analysis and machine intelligence
  (2020)

\bibitem{yang2019controllable}
Yang, S., Wang, Z., Wang, Z., Xu, N., Liu, J., Guo, Z.: Controllable artistic
  text style transfer via shape-matching gan. In: Proceedings of the IEEE/CVF
  International Conference on Computer Vision. pp. 4442--4451 (2019)

\bibitem{zhan2019spatial}
Zhan, F., Zhu, H., Lu, S.: Spatial fusion gan for image synthesis. In:
  Proceedings of the IEEE/CVF Conference on Computer Vision and Pattern
  Recognition. pp. 3653--3662 (2019)

\bibitem{zhang2018separating}
Zhang, Y., Zhang, Y., Cai, W.: Separating style and content for generalized
  style transfer. In: Proceedings of the IEEE conference on computer vision and
  pattern recognition. pp. 8447--8455 (2018)

\bibitem{zhu2017unpaired}
Zhu, J.Y., Park, T., Isola, P., Efros, A.A.: Unpaired image-to-image
  translation using cycle-consistent adversarial networks. In: Proceedings of
  the IEEE international conference on computer vision. pp. 2223--2232 (2017)

\bibitem{zhu2017multimodal}
Zhu, J.Y., Zhang, R., Pathak, D., Darrell, T., Efros, A.A., Wang, O.,
  Shechtman, E.: Multimodal image-to-image translation by enforcing bi-cycle
  consistency. In: Advances in neural information processing systems. pp.
  465--476 (2017)

\end{thebibliography}



\end{document}